\documentclass[11pt,letterpaper]{article}
\usepackage{naaclhlt2015}
\usepackage{times}
\usepackage{latexsym}
\usepackage{epsfig}
\usepackage{graphicx}
\usepackage{amsmath}
\usepackage{amssymb}
\usepackage{bm}
\usepackage[usenames]{color}
\usepackage{booktabs}
\usepackage{caption}
\usepackage{subcaption}
\usepackage{multirow}
\usepackage{footnote}
\makesavenoteenv{tabular}
\makesavenoteenv{table}

\graphicspath{{figures/}}

\DeclareMathOperator*{\argmax}{\arg\!\max}

\title{Translating Videos to Natural Language \\
Using Deep Recurrent Neural Networks}
\author{Subhashini Venugopalan \\ 
 UT Austin\\
 Austin, TX\\
 {\tt\footnotesize vsub@cs.utexas.edu}
\And Huijuan Xu  \\ 
 UMass Lowell\\
 Lowell, MA\\
 {\tt\footnotesize hxu1@cs.uml.edu}
\And Jeff Donahue \\ 
 UC Berkeley, ICSI\\
 Berkeley, CA\\
 {\tt\footnotesize jdonahue@eecs.berkeley.edu}
\AND Marcus Rohrbach\\
 UC Berkeley, ICSI\\
 Berkeley, CA\\
 {\tt\footnotesize rohrbach@eecs.berkeley.edu}
 \And  Raymond Mooney\\
 UT Austin\\
 Austin, TX\\
 {\tt\footnotesize mooney@cs.utexas.edu}
 \And Kate Saenko\\
 UMass Lowell\\
 Lowell, MA\\
 {\tt\footnotesize saenko@cs.uml.edu}
}

\date{}

\begin{document}
\maketitle

\begin{abstract}
Solving the visual symbol grounding problem has long been a goal of artificial intelligence. The field appears to be advancing closer to this goal with recent breakthroughs in deep learning for natural language grounding in static images. In this paper, we propose to translate videos directly to sentences using a unified deep neural network with both convolutional and recurrent structure. Described video datasets are scarce, and most existing methods have been applied to toy domains with a small vocabulary of possible words. By transferring knowledge from 1.2M+ images with category labels and 100,000+ images with captions, our method is able to
create sentence descriptions of open-domain videos with large vocabularies. 
We compare our approach with recent work using language generation metrics, subject, verb, and object prediction accuracy, and a human evaluation.
\end{abstract}
\section{Introduction}

\begin{figure}[t]
\footnotesize{\textit{
\textbf{Input video:}\\
   \includegraphics[width=0.31\linewidth]{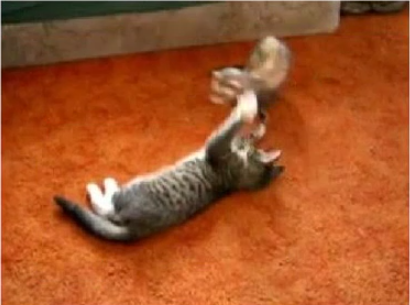}\hspace{0.05in}
   \includegraphics[width=0.31\linewidth]{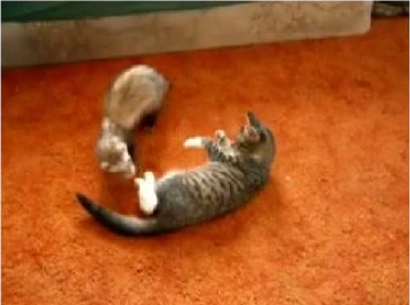}\hspace{0.05in}
   \includegraphics[width=0.31\linewidth]{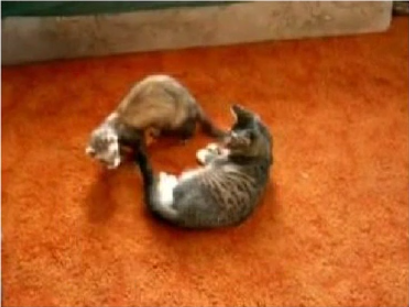} \\
\textbf{Our output:} A cat is playing with a toy. \\ 
\textbf{Humans:} 
A Ferret and cat fighting with each other. /
A cat and a ferret are playing. /
A kitten is playing with a ferret. / 
A kitten and a ferret are playfully wrestling. 
}}
\caption{
Our system takes a short video as input and outputs a natural language description of the main activity in the video.
}\label{fig:main}
\vspace{0.1cm}
\end{figure}

For most people, watching a brief video and describing what happened (in words)
is an easy task. For machines, extracting the meaning from video pixels and generating natural-sounding language is a very complex problem.
Solutions have been proposed for narrow domains with a small set of known actions and objects, e.g.,~\cite{barbu:uai12,rohrbach:iccv13}, but
generating descriptions for ``in-the-wild'' videos such as the YouTube domain (Figure~\ref{fig:main}) remains an open challenge.

Progress in open-domain video description has been difficult in
part due to large vocabularies and very limited training data
consisting of videos with associated descriptive sentences. 
Another serious obstacle has been the lack of rich models 
that can capture the joint dependencies of a
sequence of frames and a corresponding sequence of words. Previous work
has simplified the problem by detecting a fixed set of semantic
roles, such as subject, verb, and object~\cite{guadarrama:iccv13,thomason:coling14}, as an intermediate representation.
This fixed representation is problematic for large vocabularies and also leads
to oversimplified rigid sentence templates which are unable to model the
complex structures of natural language.

In this paper, we propose to translate from video pixels to natural
language with a single deep neural network. Deep NNs can learn powerful
features~\cite{donahue2013decaf,ZF14eccv}, but require a lot of supervised training data.
We address the problem  by transferring knowledge from auxiliary tasks. Each frame of the video is modeled by a convolutional (spatially-invariant) network pre-trained on 1.2M+ images with category labels~\cite{krizhevsky2012imagenet}. The meaning state and sequence of words is modeled by a recurrent (temporally invariant) deep network pre-trained on 100K+ Flickr~\cite{flickr30k} and COCO~\cite{coco2014} images with associated sentence captions. 
We show that such knowledge transfer significantly improves performance on the video task.

Our approach is inspired by recent breakthroughs reported by several research groups in image-to-text generation,
in particular, the work by~\newcite{donahueArxiv14}.
They applied a version of their model to video-to-text generation, but stopped
short of proposing an end-to-end single network, using an intermediate role representation instead. Also, they showed results only on the narrow domain of cooking videos with a small set of pre-defined objects and
actors.
Inspired by their approach, we utilize a Long-Short Term Memory
(LSTM) recurrent neural network~\cite{schmidLSTM} to model sequence
dynamics, but connect it directly to a deep convolutional neural network to
process incoming video frames,
avoiding supervised intermediate representations altogether. 
This model is similar to their image-to-text model, but we adapt it for video sequences.

Our proposed approach has several important advantages over existing
video description work. The LSTM model, which has recently achieved state-of-the-art results on machine translation tasks (French and
English~\cite{ilya:nips14}), effectively models the sequence
generation task without requiring the use of fixed sentence templates
as in previous work~\cite{guadarrama:iccv13}. Pre-training on image
and text data naturally exploits related data to supplement the limited
amount of descriptive video currently available.
Finally, the deep convnet, the winner of
the ILSVRC2012~\cite{imagenet2014} image classification competition, provides a strong
visual representation of objects, actions and scenes depicted in the
video. 

Our main contributions are as follows: 
\begin{itemize}
\vspace{-0.1in}
\item We present the first end-to-end deep model for video-to-text generation that
  simultaneously learns  a
  latent ``meaning'' state, and a fluent grammatical model of the associated
  language.
\vspace{-0.2in}
\item We leverage still image classification and caption data and transfer deep networks learned on such data to the video domain.
\vspace{-0.1in}
\item We provide a detailed evaluation of our model on the popular YouTube corpus~\cite{chen:acl11} and demonstrate a significant improvement over the state of the art.
\end{itemize}

\section{Related Work}

Most of the existing research in video description has focused on narrow
domains with limited vocabularies of objects and
activities~\cite{kojima,lee2008save,khan:eacl-wkshp12,barbu:uai12,ding2012beyond,khan:eacl-wkshp12,das2013thousand,DaSrCoWSDM2013,rohrbach:iccv13,yu2013grounded}.
For example, \newcite{rohrbach:iccv13},~\newcite{rohrbach14gcpr} produce descriptions for videos of several people cooking in the same kitchen.
These approaches generate sentences
by first predicting a semantic role
representation, e.g., modeled with a CRF, of high-level concepts such as the actor, action and object. Then they use a template or statistical machine translation to translate
the semantic representation to a sentence. 

Most work on ``in-the-wild'' online video has focused on retrieval and predicting event tags rather than generating descriptive sentences; examples are tagging YouTube~\cite{google} and retrieving online video in the TRECVID competition~\cite{over12tv}. Work on TRECVID has also included clustering both video and text features for video retrieval, e.g.,~\cite{wei2010multimodal,huang2013multi}.

The previous work on the YouTube corpus we employ
\cite{motwani:ecai12,krishnamoorthy:aaai13,guadarrama:iccv13,thomason:coling14}
used a two-step approach, first detecting a fixed tuple of role
words, such as subject, verb, object, and scene, and then using a
template to generate a grammatical sentence. They also utilize
language models learned from large text corpora to aid visual
interpretation as well as sentence generation. We compare our method to the
best-performing method of~\newcite{thomason:coling14}.
A recent paper by~\newcite{corsoAAAI} extracts deep features from video
and a continuous vector from language, and projects both to a joint semantic space.
They apply their joint embedding to SVO prediction and generation, but do not
provide quantitative generation results. Our network learns a joint state
vector implicitly, and additionally models sequence dynamics of the language.

Predicting natural language desriptions of still images has
 received considerable attention, with some of the earliest works
 by~\newcite{aker2010acl},~\newcite{farhadi2010every},~\newcite{yao2010i2t}, and
 \newcite{kulkarni2011baby} amongst others.
Propelled by successes of deep learning, several groups released record breaking 
results in just the past year~\cite{donahueArxiv14,yuille14,karpathy14nips,fangMSRArxiv14,kirosArxiv14,vinyalsArxiv14,kuznetsova2014treetalk}.

In this work, we use deep recurrent nets (RNNs), which have recently
demonstrated 
strong results for machine translation tasks using Long Short Term
Memory (LSTM) RNNs~\cite{ilya:nips14,cho2014properties}.
In contrast to traditional statistical MT~\cite{koehn10book}, 
RNNs naturally combine
with vector-based representations, such as those for images and video.
\newcite{donahueArxiv14} and \newcite{vinyalsArxiv14} simultaneously proposed a multimodal analog
of this model, with an architecture which uses a visual convnet to
encode a deep state vector, and an LSTM to decode the vector into a
sentence.  

Our approach to video to text generation is inspired by the work of~\newcite{donahueArxiv14}, who also applied a variant of their model to
video-to-text generation, but stopped short of training an end-to-end
model. Instead they converted the video to an intermediate role representation using 
a CRF, then decoded that representation into a sentence. In contrast,
we bypass detection of high-level roles and use the output of a deep
convolutional network directly as the state vector that is decoded into a
sentence.  
This avoids the need for labeling semantic roles, which can be difficult to detect in the case of very large vocabularies. It also allows us to
first pre-train the model on a large image and caption database, and transfer the 
knowledge to the video domain where the corpus size is smaller.
While~\newcite{donahueArxiv14} only showed results on a narrow domain of cooking videos with a small set of pre-defined objects and actors, we generate sentences for open-domain YouTube videos with a vocabulary of thousands of words.

\section{Approach}
\label{sec:approach}

\begin{figure}[t]
\begin{center}
   \includegraphics[width=\linewidth]{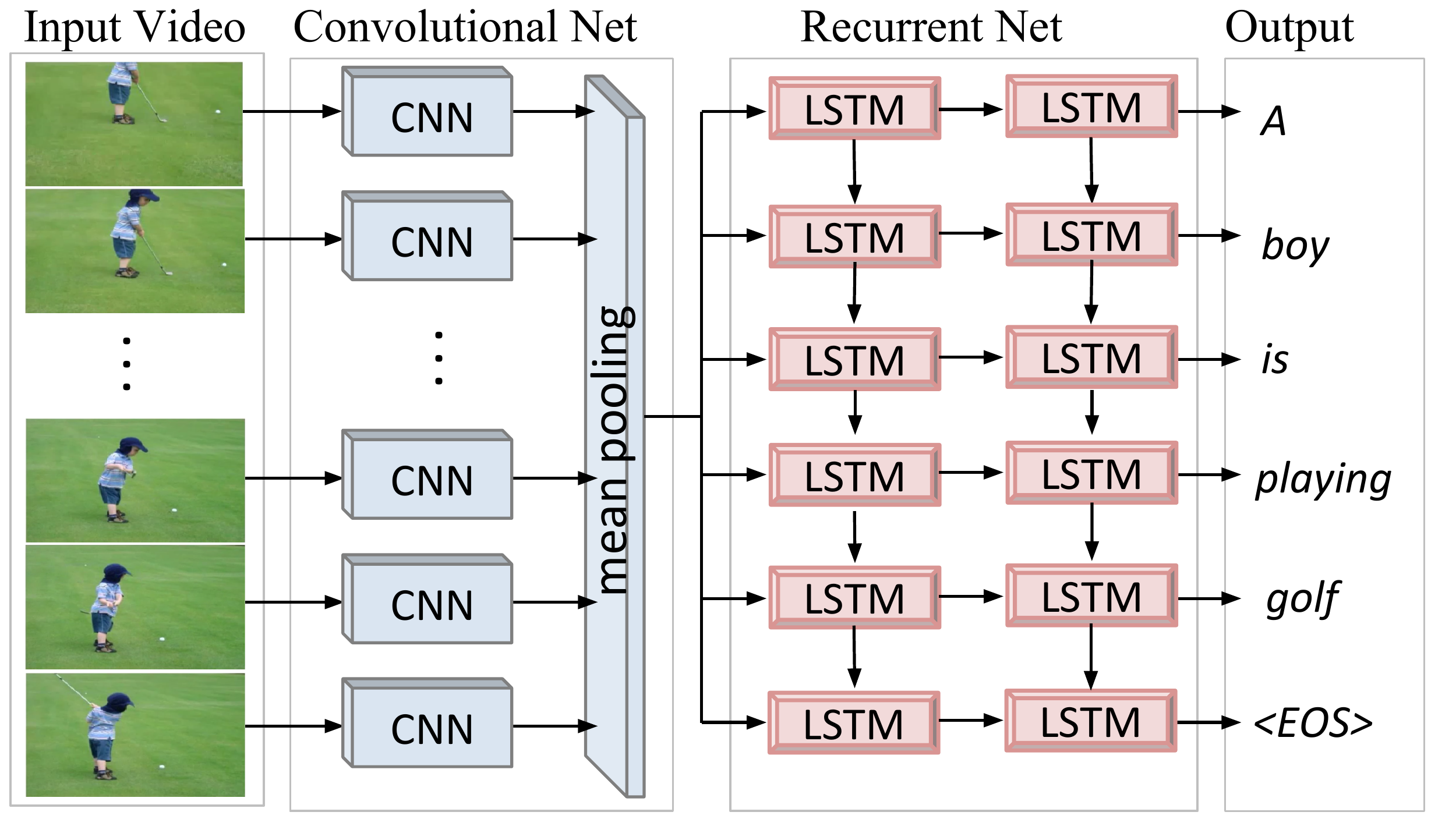}
\caption{The structure of our video description network. We extract fc$_7$
features for each frame, mean pool the features across the entire video and
input this at every time step to the LSTM network. The LSTM outputs one word at
each time step, based on the video features (and the previous word) until it
picks the end-of-sentence tag.}
\label{fig:network}
\end{center}
\end{figure}

Figure~\ref{fig:network} depicts our model for sentence generation
from videos. Our framework is based on deep image description
models in \newcite{donahueArxiv14};\newcite{vinyalsArxiv14} and extends them to generate sentences
describing events in videos. These models work by first applying a
feature transformation on an image to generate a fixed dimensional
vector representation. They then use a sequence model, specifically a
Recurrent Neural Network (RNN), to ``decode'' the vector into a
sentence (i.e. a sequence of words). In this work, we apply the same
principle of ``translating'' a visual vector into an English sentence
and show that it works well for describing dynamic videos as well as
static images.

We identify the most likely description for a given video by training
a model to maximize the log likelihood of the sentence $S$,
given the corresponding video $V$ and the model parameters $\theta$,
\vspace{-0.1in}
\begin{equation}\label{eqn:loglikelihood}
 \theta^* = \argmax_{\theta} \text{ } \displaystyle\sum_{(V,S)} \text{log }p(S | V; \theta) 
 \end{equation}
 \vspace{-0.2in}

Assuming a generative model of $S$ that produces each word in the
sequence in order, the log probability of the sentence is given by
the sum of the log probabilities over the words and can be expressed
as:
\vspace{-0.2in}
$$ \text{ log } p(S | V) = \displaystyle\sum_{t=0}^{N} \text{log }p(S_{w_t} | V,
S_{w_1}, \ldots, S_{w_{t-1}} )$$
where $S_{w_i}$ represents the $i^{th}$ word in the sentence and N is the total
number of words. Note that we have dropped $\theta$ for convenience. 

A sequence model would be apt to model $p(S_{w_t} | V,
S_{w_1}, \ldots, S_{w_{t-1}} )$, and we choose an RNN. An RNN, parameterized by $\theta$,
maps an input $x_t$, and the previously seen words expressed as a hidden state
or memory, $h_{t-1}$ to an output $z_t$ and an updated state $h_t$ using a
non-linear function $f$:
\begin{equation}
h_t = f_{\theta}(x_t, h_{t-1})
\end{equation}
where $(h_0 = 0)$. In our work we use the highly successful Long Short-Term Memory (LSTM) net as
the sequence model, since it has shown superior performance on tasks
such as speech recognition \cite{graves2014towards}, machine translation
\cite{ilya:nips14,cho2014properties} and the more related task of generating
sentence descriptions of images \cite{donahueArxiv14,vinyalsArxiv14}. To be specific, we use two layers of
LSTMs (one LSTM stacked atop another) as shown in Figure \ref{fig:network}.
We present details of the network in Section \ref{subsec:lstm}.
To convert videos to a fixed length representation (input $x_t$), we
use a Convolutional Neural Network (CNN). 
 We present details
of how we apply the CNN model to videos in Section
\ref{subsec:vidrep}.

\subsection{LSTMs for sequence generation} \label{subsec:lstm}

A Recurrent Neural Network (RNN) is a generalization of feed forward neural
networks to sequences. Standard RNNs learn to map a sequence of inputs
($x_1,\ldots, x_t)$ to a sequence of hidden states ($h_1,\ldots, h_t)$, and from
the hidden states to a sequence of outputs ($z_1, \ldots, z_t$) based on the
following recurrences:
\begin{align}
h_t &= f (W_{xh} x_t + W_{hh} h_{t-1}) \\
z_t &= g (W_{zh} h_t)
\end{align}
where $f$ and $g$ are element-wise non-linear functions such as a
sigmoid or hyperbolic tangent, $x_t$ is a fixed length vector
representation of the input, $h_t \in \mathbb{R}^N$ is the hidden
state with $N$ units, $W_{ij}$ are the weights connecting the layers
of neurons, and $z_t$ the output vector.

\begin{figure}
\centering
            \includegraphics[width=0.5\textwidth]{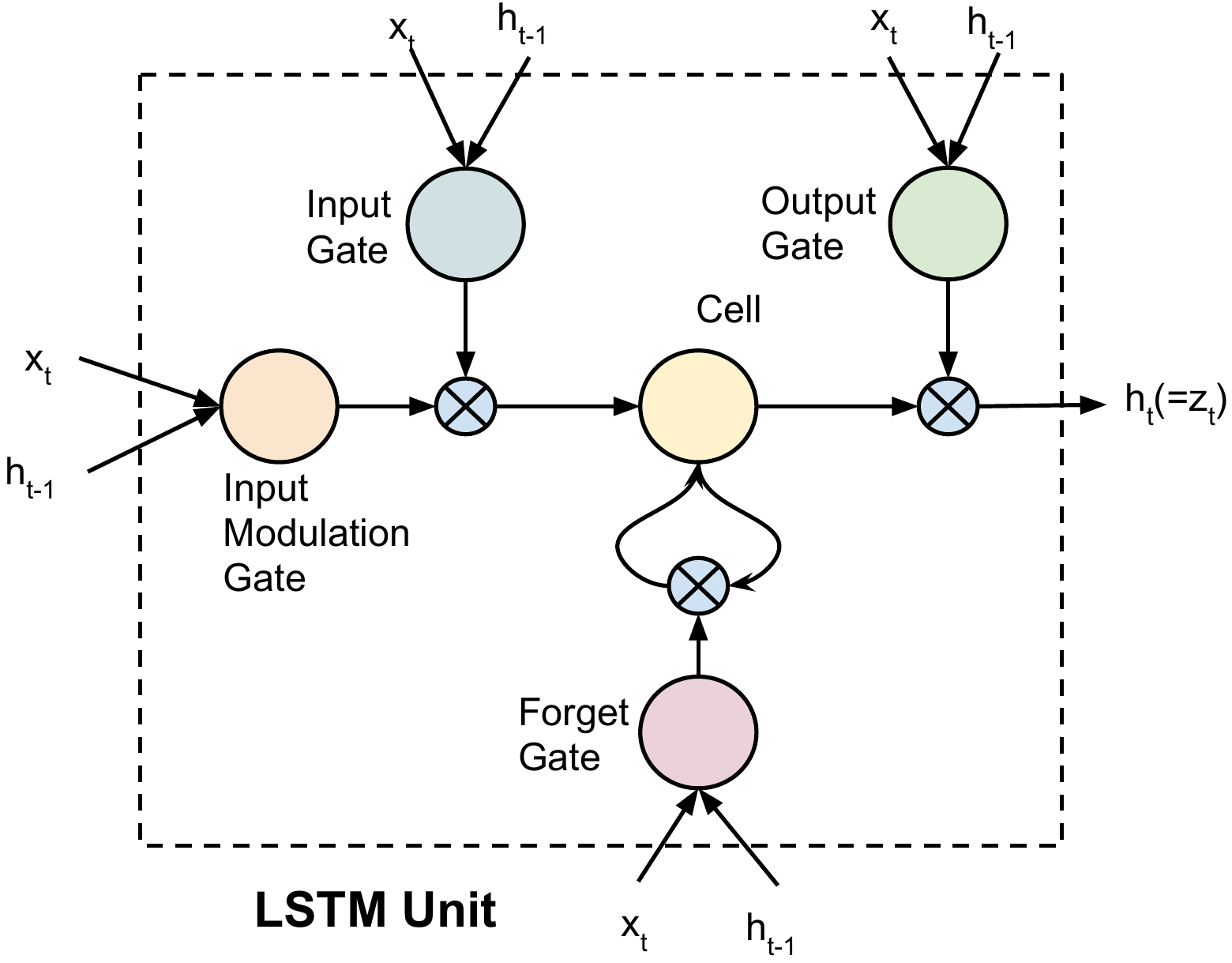}
    \caption{The LSTM unit replicated from \cite{donahueArxiv14}.
The memory cell is at the core of the LSTM unit and it is modulated by the 
input, output and forget gates controlling how much knowledge is transferred
 at each time step.}
    \label{fig:lstm_unit}
\vspace{0.1cm}
\end{figure}

RNNs can learn to map sequences for which the alignment between the
inputs and outputs is known ahead of time \cite{ilya:nips14} however it's
unclear if they can be applied to problems where the inputs ($x_i$)
and outputs ($z_i$) are of varying lengths. This problem is solved by
learning to map sequences of inputs to a fixed length vector using one
RNN, and then map the vector to an output sequence using another RNN.
Another known problem with RNNs is that, it can be difficult to train
them to learn long-range dependencies
\cite{hochreiter2001gradient}. However, LSTMs \cite{schmidLSTM}, which
incorporate explicitly controllable memory units, are known to be able
to learn long-range temporal dependencies.
In our work we use the LSTM unit
in Figure \ref{fig:lstm_unit},
described in \newcite{zaremba2014learning}, and \newcite{donahueArxiv14}.

At the core of the LSTM model is a memory cell $c$ which encodes, at every time step,
the knowledge of the inputs that have been observed up to that step.
The cell is modulated by gates 
which are all sigmoidal, having range $[0,1]$, and are applied multiplicatively. 
The gates determine whether the LSTM keeps the value from the
gate (if the layer evaluates to 1) or discards it (if it evaluates to 0). The three gates --
input gate ($i$) controlling whether the LSTM considers its current input ($x_t$),
the forget gate ($f$) allowing the LSTM to forget its previous memory ($c_{t-1}$),
 and the output gate ($o$)
deciding how much of the memory to transfer to the hidden state ($h_t$), all enable the LSTM to learn
complex long-term dependencies. The recurrences for the LSTM are then 
defined as:
\vspace{-0.1in}
\begin{align}
i_t &= \sigma(W_{xi} x_t + W_{hi} h_{t-1}) \\
f_t &= \sigma(W_{xf} x_t + W_{hf} h_{t-1}) \\
o_t &= \sigma(W_{xo} x_t + W_{ho} h_{t-1}) \\
c_t &=  f_t \odot c_{t-1}  + i_t \odot \phi(W_{xc} x_t + W_{hc} h_{t-1}) \\
h_t &= o_t \odot \phi(c_t)
\vspace{-0.2in}
\end{align}
where $\sigma$ is the sigmoidal non-linearity, %
$\phi$ is the hyperbolic tangent non-linearity, $\odot$ represents the product
with the gate value, and the weight matrices denoted by $W_{ij}$ are the trained parameters. 

\subsection{CNN-LSTMs for video description}\label{subsec:vidrep}
We use a two layer LSTM model for generating descriptions
for videos 
based on experiments by \newcite{donahueArxiv14} which suggest two LSTM layers
are better than four and a single layer for image to text tasks.
We employ the LSTM to ``decode" a visual
feature vector representing the video to generate textual output. The
first step in this process is to generate a fixed-length visual input
that effectively summarizes a short video. For this we use a CNN,
specifically the publicly available
\textit{Caffe} \cite{caffe} reference model, a minor variant of
\textit{AlexNet} \cite{krizhevsky2012imagenet}. 
The net is pre-trained on the 1.2M image
ILSVRC-2012 object classification subset of the ImageNet dataset
\cite{imagenet2014} and hence provides a robust initialization for
recognizing objects and thereby expedites training.
We sample frames in
the video (1 in every 10 frames) and extract the output of the fc$_7$ layer 
and perform a mean pooling over the frames to
generate a single 4,096 dimension vector for each video. The resulting
visual feature vector forms the input to the first LSTM layer. We
stack another LSTM layer on top as in Figure~\ref{fig:network}, and the hidden
state of the LSTM in the first layer is the input to the LSTM unit in
the second layer. A word from the sentence forms the target of the
output LSTM unit. In this work, we represent words using 
``one-hot" vectors (i.e 1-of-N coding, where is N is the vocabulary size).

\paragraph{Training and Inference:}
The two-layer LSTM model is trained to predict the next word $S_{w_t}$ in the sentence
given the visual features and the previous $t-1$ words, $p(S_{w_t} | V,
S_{w_1}, \ldots, S_{w_{t-1}} )$. During training the visual feature,
sentence pair ($V,S$) is provided to the model, which then optimizes the
log-likelihood (Equation \ref{eqn:loglikelihood})
over the entire training dataset using
stochastic gradient descent. At each time step,
the input %
$x_t$ is fed to the LSTM along with the previous time step's hidden state
$h_{t-1}$ and the LSTM emits the next hidden state vector $h_t$ (and a word).
For the first layer of the LSTM $x_t$ is the concatenation of the visual feature vector and the previous encoded word ($S_{w_{t-1}}$, the ground truth word during training and the predicted word during test time). For the second layer of the LSTM $x_t$ is $z_t$ of the first layer.
Accordingly, inference must also be performed sequentially in the
order $h_1 = f_W(x_1, 0)$, $h_2 = f_W(x_2, h_1)$, until the model emits the
end-of-sentence (EOS) token
at the final step $T$. In our model the output ($h_t$ = $z_t$) of the second
layer LSTM unit is used to obtain the emitted word. We apply the Softmax
function, to get a probability distribution over the words $w$ in the vocabulary
$D$. 
\begin{equation} 
p(w|z_t) = \frac{\text{exp}(W_{w}z_{t})}{\sum_{w' \in
D} \text{exp}(W_{w'}z_{t})}
\end{equation}  
where %
$W_w$ is 
a learnt embedding vector for word $w$. 
At test time, we choose the word $\hat{w}$ with the maximum probability for each time step $t$ until we obtain the EOS token.

\subsection{Transfer Learning from Captioned Images}\label{subsec:transfer}
Since the training data available for video description is quite
limited (described in Section \ref{subsec:data}), we also leverage much larger
datasets available for image captioning to train our LSTM model and
then fine tune it on the video dataset. Our LSTM model
for images is the same as the one described above for single video
frames (in Section \ref{subsec:lstm}, and \ref{subsec:vidrep}). As
with videos, we extract fc$_7$ layer features (4096
dimensional vector) from the network (Section \ref{subsec:vidrep}) for the images. This forms the visual feature that
is input to the  2-layer LSTM description model. 
The vocabulary is the combined set of words in the video and image datasets.
After the model is
trained on the image dataset, we use the weights of the trained model
to initialize the LSTM model for the video description
task. Additionally, we reduce the learning rate on our LSTM model to
allow it to tune to the video dataset. This speeds up training
and allows exploiting knowledge previously learned for image description.

\section{Experiments}

\subsection{Datasets}\label{subsec:data}
\paragraph{Video dataset.} We perform all our experiments on the Microsoft
Research Video Description Corpus \cite{chen:acl11}. This video corpus is a
collection of 1970 YouTube snippets. The duration of each clip is between
10 seconds to 25 seconds, typically depicting a single activity or a short
sequence. The dataset comes with several human generated
descriptions in a number of languages; we use the roughly 40 available English
descriptions per video.
This dataset (or portions of it) have been
used in several prior works
\cite{motwani:ecai12,krishnamoorthy:aaai13,guadarrama:iccv13,thomason:coling14,corsoAAAI}
on action recognition and video description tasks.
For our task we pick 1200 videos to be used as training data, 100 videos for validation
 and 670 videos for testing, 
as used
by the prior works on video description
\cite{guadarrama:iccv13,thomason:coling14,corsoAAAI}.

\paragraph{Domain adaptation, image description datasets.}
Since the number of videos for the description task is quite small when compared to the size of
the datasets used by LSTM models in other tasks such as translation
\cite{ilya:nips14} (12M sentences),
we use data from the Flickr30k and COCO2014 datasets for training
 and learn to adapt to the video dataset by fine-tuning the image description models.
 The Flickr30k \cite{flickr30k} dataset has about 30,000 images, each with 5 or more
 descriptions. We hold out 1000 images at random for validation and use the remaining for training.
 In addition to this, we use the recent COCO2014 \cite{coco2014} image
 description dataset consisting of 82,783 training images and 40,504 validation
 images, each with 5 or more sentence descriptions. We perform ablation
 experiments by training models on each dataset individually, and on the
 combination and report results on the YouTube video test dataset.

\subsection{Models}
\paragraph{HVC} This is the Highest Vision Confidence model described in
\cite{thomason:coling14}. The model uses strong visual detectors to predict
confidence over 45 subjects, 218 verbs and 241 objects. 
\paragraph{FGM} \cite{thomason:coling14}
also propose a factor graph model (FGM) that combines knowledge mined from text
corpora with visual confidences from the HVC model using a factor graph and
performs probabilistic inference to determine the most likely subject, verb, object and scene
tuple. They then use a simple template to generate a sentence from the tuple. 
In this work, we compare the output of our model to the subject, verb, object words predicted
by the HVC and FGM models and the sentences generated from the SVO triple.

\paragraph{Our LSTM models} We present four main models. LSTM-YT is our base
two-layer LSTM model %
trained on the YouTube video dataset. LSTM-YT$_{flickr}$ is
the model trained on the Flickr30k \cite{flickr30k} dataset, and fine
tuned on the YouTube dataset as descibed in Section \ref{subsec:transfer}.
LSTM-YT$_{coco}$ is first trained on the COCO2014 \cite{coco2014} dataset and then
fine-tuned on the video dataset. Our final model, LSTM-YT$_{cocoflickr}$ is 
trained on the combined data of both the
Flickr and COCO models and is tuned on YouTube. 
To compare the overlap in
content between the image dataset and YouTube dataset, we use the
model trained on just the Flickr images (LSTM$_{flickr}$) and just the COCO
images (LSTM$_{coco}$) and evaluate their performance on the test videos.

\subsection{Evaluation Metrics and Results}
\label{subsec:evalmetrics}

\paragraph{SVO accuracy.} Earlier
works \cite{krishnamoorthy:aaai13,guadarrama:iccv13} that reported results on the
YouTube dataset compared their method based on how well their model could
predict the subject, verb, and object (SVO) depicted in the video.
 Since these models first predicted the content (SVO triples) and then
 generated the sentences, the S,V,O accuracy captured the quality of the content
 generated by the models. However, in our case the sequential LSTM directly
 outputs the sentence, %
 so we extract the S,V,O from the dependency parse of the generated
 sentence. 
We present, in Table \ref{tab:binAnyValidSVO} and Table
\ref{tab:binMostFreqSVO},
the accuracy of S,V,O words comparing the performance of our model
against any valid ground truth triple and the most frequent triple found in
human description for each video.
The latter evaluation was also reported by~\cite{corsoAAAI}, so we include it 
here for comparison.

\begin{table}[h]
\begin{center}
\begin{tabular}{l|c|c|c} 
\toprule
Model &	{\bf{S\%}} & {\bf{V\%}} & {\bf{O\%}} \\ \hline
HVC {\tiny{\cite{thomason:coling14}}} & 86.87& 38.66&  22.09 \\ 
FGM {\tiny{\cite{thomason:coling14}}} & \bf{88.27}& 37.16&  24.63 \\ \hline
LSTM$_{flickr}$ & 79.95 & 15.47 & 13.94 \\
LSTM$_{coco}$ & 56.30 & 06.90 & 14.86 \\
LSTM-YT  & 79.40 & 35.52& 20.59\\ 
LSTM-YT$_{flickr}$  & 84.92 & 38.66 & 21.64 \\ 
LSTM-YT$_{coco}$  & 86.58 & 42.23& \bf{26.69} \\ 
LSTM-YT$_{coco+flickr}$ & 87.27 & \bf{42.79} & 24.23 \\ \hline
\end{tabular}
\caption{SVO accuracy: Binary SVO accuracy compared against any valid S,V,O
triples in the ground truth descriptions. We extract S,V,O values from
sentences output by our model using a dependency parser. The model is
correct if it identifies S,V, or O mentioned in any one of the multiple human
descriptions.}
\label{tab:binAnyValidSVO}
\end{center}
\end{table}

\begin{table}[h]
\begin{center}
\begin{tabular}{l|c|c|c} 
\toprule
Model &	{\bf{S\%}} & {\bf{V\%}} & {\bf{O\%}} \\ \hline
HVC {\tiny{\cite{thomason:coling14}}}  & {76.57}& 22.24& 11.94 \\ 
FGM {\tiny{\cite{thomason:coling14}}}  & 76.42& 21.34& 12.39 \\ 
JointEmbed\footnotemark%
 {\tiny\cite{corsoAAAI}} & \bf{78.25} & 24.45 & 11.95 \\ \hline
LSTM$_{flickr}$ & 70.80 & 10.02 & 07.84 \\ 
LSTM$_{coco}$ & 47.44 & 02.85 & 07.05 \\
LSTM-YT  & 71.19 & 19.40 & 09.70 \\ 
LSTM-YT$_{flickr}$ & 75.37 & 21.94& 10.74 \\ 
LSTM-YT$_{coco}$ & 76.01 & 23.38 & \bf{14.03} \\ 
LSTM-YT$_{coco+flickr}$ & 75.61 & \bf{25.31} & 12.42 \\ \hline
\end{tabular}
\caption{SVO accuracy: Binary SVO accuracy compared against most frequent S,V,O
triple in the ground truth descriptions. We extract S,V,O values from
parses of sentences output by our model using a dependency parser. The model 
is correct only if it outputs the most frequently mentioned S, V, O among the
human descriptions.
}
\label{tab:binMostFreqSVO}
\end{center}
\end{table}

\paragraph{Sentence Generation.}

To evaluate the generated sentences we use the BLEU~\cite{papineni2002bleu} and METEOR~\cite{banerjee2005meteor} scores against all ground truth sentences.
BLEU is the metric that is seen more commonly in image description literature,
but a more recent study \cite{elliott2014comparing} has shown
METEOR to be a better evaluation metric. However, since both 
metrics have been shown to correlate well with human evaluations, we compare the 
generated sentences using both and present our results in Table \ref{tab:bleumeteor}.
\footnotetext{They evaluate against a filtered set of groundtruth SVO
words which provides a tiny boost to their scores.}

\begin{table}
\begin{center}
\begin{tabular}{l|cc}
\toprule
Model & BLEU & METEOR \\
\midrule
FGM {\tiny{\cite{thomason:coling14}}}& 13.68 & 23.90 \\ 
LSTM-YT  & 31.19 & 26.87 \\ 
LSTM-YT$_{flickr}$ & 32.03 & 27.87 \\ 
LSTM-YT$_{coco}$  & \bf{33.29} & \bf{29.07} \\ 
LSTM-YT$_{coco+flickr}$& \bf{33.29} & 28.88 \\ \hline
\end{tabular}
\end{center}
\caption{ 
Scores for BLEU at 4 (combined n-gram 1-4), and METEOR scores
from automated evaluation metrics comparing the quality of the
generation.
All values are reported as percentage (\%).
}
\label{tab:bleumeteor}
\end{table}

\paragraph{Human Evaluation.}
We used Amazon Mechanical Turk to also collect human judgements. We created a task
which employed three Turk workers to watch each video, and rank sentences generated by
the different models from ``Most Relevant" (5) to ``Least Relevant" (1).
No two sentences could have the same rank unless they were identical.
We also evaluate sentences on grammatical correctness. We created a different
task which required workers to rate sentences based on grammar. This task
displayed only the sentences and did not show any video. Here, workers had
to choose a rating between 1-5 for each sentence. Multiple sentences could have
the same rating. We discard responses from
workers who fail gold-standard items
and report the mean ranking/rating for each of the evaluated models in Table
\ref{tab:humaneval}.

\begin{table}[t]
\begin{center}
\begin{tabular}{lcc}
\toprule
Model & Relevance & Grammar \\
\midrule
FGM {\tiny{\cite{thomason:coling14}}} & 2.26&  \bf{3.99} \\ 
LSTM-YT  & 2.74 & 3.84  \\ 
LSTM-YT$_{coco}$  & \bf{2.93} & 3.46   \\ 
LSTM-YT$_{coco+flickr}$ & 2.83 & 3.64   \\ \hline
GroundTruth	&	4.65 & 4.61	\\ \hline
\end{tabular}
\end{center}
\caption{
Human evaluation mean scores. Sentences were uniquely ranked between 1 to 5
based on their relevance to a given video. Sentences were rated between 1 to 5
for grammatical correctness. Higher values are better.
}
\vspace{0.1cm}
\label{tab:humaneval}
\end{table}

\paragraph{Individual Frames.}\label{para:indiframe}
In order to evaluate the effectiveness of mean pooling, we performed experiments
to train and test the model on individual frames from the video. 
Our first set
of experiments involved testing how well the image description models performed on a randomly
sampled frame in the video. Similar to Tables \ref{tab:binAnyValidSVO} and
\ref{tab:binMostFreqSVO}, the model trained on Flickr30k when tested on random
frames from the video scored better on 
subjects and verbs with any valid accuracy of $75.16\%$ and $11.65\%$
respectively; and $9.01\%$ on objects. The one
trained on COCO did better on objects ($12.54\%$, any valid accuracy) but very poorly on subjects
and verbs.
In our next experiment, we used image description models (trained on Flickr30k, COCO or a
combination of both) and fine-tuned them on individual frames in the video by
picking a different frame for each description in the YouTube dataset. These
models were tested on a random frame from the test video. The overall trends in the results
were similar to those seen in Tables \ref{tab:binAnyValidSVO} and
\ref{tab:binMostFreqSVO}.
The model trained on COCO and fine-tuned on individual
video frames performed best with any valid S,V,O accuracies
$84.8\%, 38.98\%, \text{ and } 22.34\%$ respectively. The one trained on both
COCO and Flickr30k had any valid S,V,O accuracies of $85.67\%, 38.83\%, \text{
    and } 19.72\%$. We report the generation results for these models
in Table \ref{tab:bleumeteorframe}.
\begin{table}
\begin{center}
\begin{tabular}{l|cc}
\toprule
Model (individual frames) & BLEU & METEOR \\
\midrule
LSTM$_{flickr}$ & 08.62 & 18.56 \\ 
LSTM$_{coco}$ & 11.39 & 20.03 \\ 
LSTM-YT-frame$_{flickr}$ & 26.75 & 26.51 \\ 
LSTM-YT-frame$_{coco}$  & \bf{30.77} & \bf{27.66} \\ 
LSTM-YT-frame$_{coco+flickr}$& 29.72 & 27.65 \\ \hline
\end{tabular}
\end{center}
\caption{ 
Scores for BLEU at 4 (combined n-gram 1-4), and METEOR scores
comparing the quality of sentence generation by the models trained on Flickr30k
and COCO and tested on a random frame from the video. 
LSTM-YT-frame models were fine tuned on individual frames from the Youtube video
dataset.
All values are reported as percentage (\%).
}
\vspace{0.1cm}
\label{tab:bleumeteorframe}
\end{table}

\section{Discussion}

\textbf{Image only models.} The models trained purely on the image description
data LSTM$_{flickr}$ and LSTM$_{coco}$ achieve lower accuracy on the verbs
and objects (Tables \ref{tab:binAnyValidSVO}, \ref{tab:binMostFreqSVO})
since the YouTube videos encompass a wider domain and a variety of
actions not detectable from static images.

\textbf{Base LSTM model.} We note that in the SVO binary accuracy metrics 
(Tables \ref{tab:binAnyValidSVO} and \ref{tab:binMostFreqSVO}),
the base LSTM model (LSTM-YT) achieves a slightly lower accuracy compared to prior work.
This is likely due to the fact that previous work explicitly 
optimizes to identify the best subject, verb and object for a video; whereas the
LSTM model is trained on objects and actions jointly in a sentence and needs to 
learn to interpret these in different contexts. However, with regard to the
generation metrics BLEU and METEOR, training based on the full sentence helps
the LSTM model develop fluency and vocabulary similar to that seen in the
training descriptions and
allows it to outperform the template based generation.

\begin{figure}[!htb]
\begin{center}
\vspace{-0.2cm}
  \def\svgwidth{0.5\textwidth} 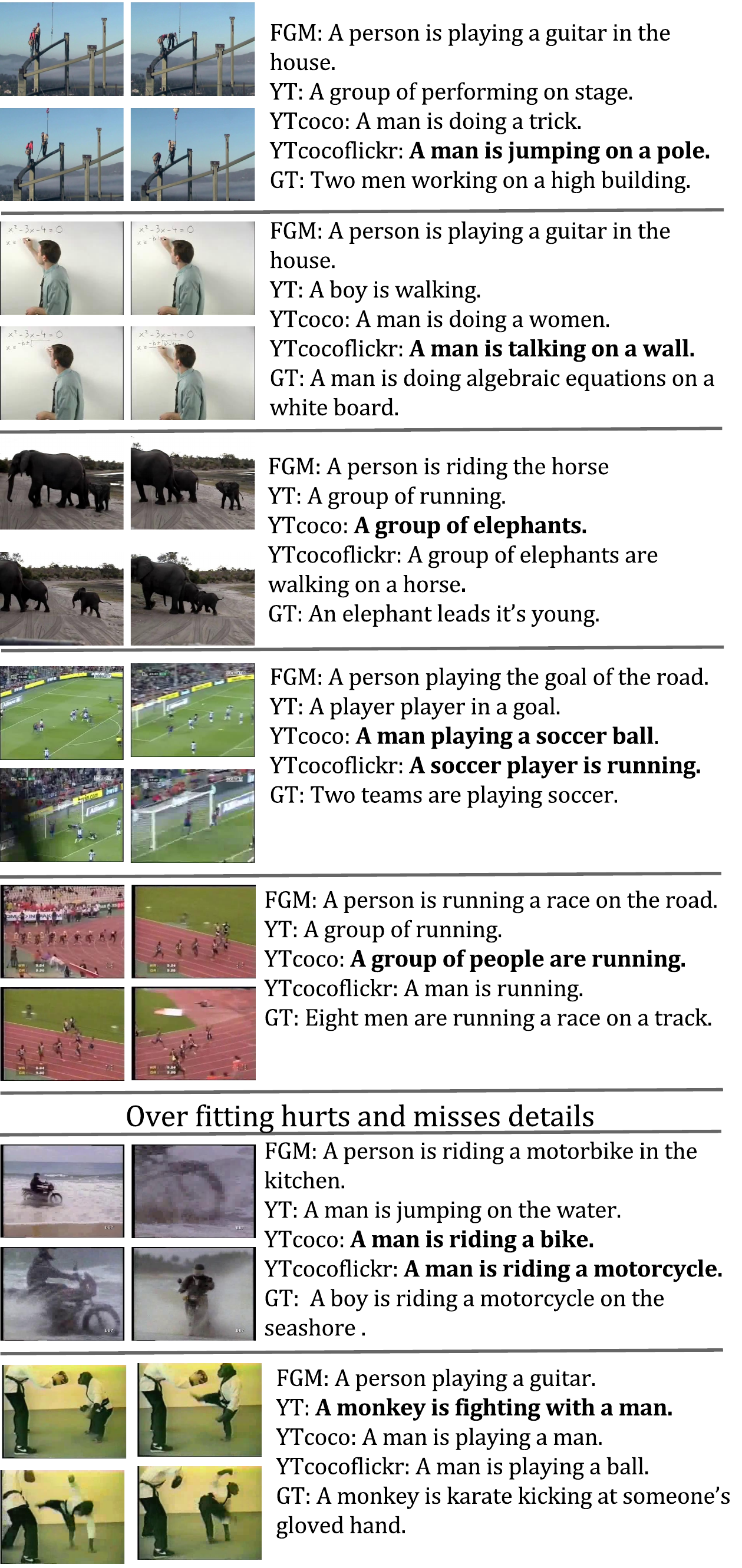
  \caption{Examples to demonstrate effectiveness of transferring from the
  image description domain. YT refer to the LSTM-YT, YTcoco to the
  LSTM-YT$_{coco}$, and YTcocoflickr to the LSTM-YT$_{coco+flickr}$ models. GT is a random
  human description in the ground truth. 
  Sentences in \textbf{bold} highlight the most accurate description for the
  video amongst the models.
  Bottom two examples show how transfer
  can overfit. Thus, while base LSTM-YT model detects water and monkey, 
  the LSTM-YT$_{coco}$ and LSTM-YT$_{cocoflickr}$ models fail to describe the event completely.
  }
  \label{fig:cherry_lemons}
\end{center}
\end{figure}

\textbf{Transferring helps.} 
From our experiments, it is clear that learning from the image description data
improves the performance of the model in all criteria of evaluation. We present
a few examples demonstrating this in Figure \ref{fig:cherry_lemons}.
The model that was pre-trained on COCO2014 shows a larger performance
improvement, indicating that our model can effectively leverage a large
auxiliary source of training data to improve its object and verb predictions.
The model pre-trained on the combined data of Flickr30k and
COCO2014 shows only a marginal improvement,
perhaps due to overfitting. Adding dropout as in~\cite{vinyalsArxiv14} 
is likely to help prevent overfitting and improve
performance.

From the automated evaluation in Table \ref{tab:bleumeteor}
it is clear that the fully deep
video-to-text generation models outperform previous work. As mentioned
previously, training on the full sentences is probably the main reason for the
improvements.

\textbf{Testing on individual frames.}
The experiments that evaluated models on individual frames
(Section \ref{para:indiframe})
from the video have
trends similar to those seen on mean pooled frame features. Specifically, the model
trained on Flickr30k, when directly evaluated on YouTube video frames performs
better on subjects and verbs, whereas the one trained on COCO does better on
objects. This is explained by the fact that Flickr30k images are more varied but
COCO has more examples of a smaller collection of objects, thus increasing
object accuracy. 
Amongst the models trained on images and individual video frames,
the ones trained on COCO (and the combination of both) perform well, but are
still a bit poorer compared to the models trained on mean-pooled features. 
One point to note however is that, these models were trained and evaluated on
random frames from the video, and not necessarily a key-frame or
most-representative frame. It's likely that choosing a representative frame from
the video might result in a small improvement. But, on the whole, our experiments
show that models trained on images alone do not directly perform well on video
frames, and a better representation is required to learn from videos.

\textbf{Mean pooling is significant.}
Our additional experiments that trained and tested on individual frames in the
video, reported in section \ref{subsec:evalmetrics}, suggest that mean pooling 
frame features gives significantly better results. 
This could potentially indicate that mean pooling features 
across all frames in the video is a reasonable representation for short video clips
at least for the task of generating simple sentential descriptions.

\textbf{Human evaluation.} We note that the sentences generated by our model
have been ranked more relevant (Table \ref{tab:humaneval}) to the
content in the video than previous models.
 However, there is still a significant gap between the human ground truth
sentence and the ones generated by the LSTM models.
Additionally, when we ask Turkers to rate only the sentences (they are not
provided the video) on grammatical correctness,
the template based FGM \cite{thomason:coling14} achieves the highest ratings.
This can be explained by the fact that their work uses a 
template technique to
generate sentences from content, and is hence grammatically well formed. Our
model sometimes predicts prepositions and articles more frequently, resulting in
duplicates and hence incorrect grammar.
%

%We present some sample videos and descriptions where pre-training on the image
%data improves the performance our models.
\vspace{-0.2cm}

\section{Conclusion}
In this paper we have proposed a model for video description which uses neural
networks for the entire pipeline from pixels to sentences
 and can potentially allow for the training and tuning of the entire network.
In an extensive experimental evaluation, we
showed that our approach generates better sentences than related approaches.
We also showed that exploiting image description data improves performance
compared to relying only on video description data. 
However our approach
falls short in better utilizing the temporal information in videos,
which is a good direction for future work.
We will release our \textit{Caffe}-based implementation, as well as the model and generated sentences.

\section*{Acknowledgments}
The authors thank Trevor Darrell for his valuable advice. We would also like to
thank reviewers for their comments and suggestions. 
Marcus Rohrbach was supported by a fellowship within the FITweltweit-Program of
the German Academic Exchange Service (DAAD).
This research was partially supported by ONR ATL Grant N00014-11-1-010, NSF
Awards  IIS-1451244 and IIS-1212798.

\clearpage
\bibliographystyle{naaclhlt2015}
\bibliography{youtube2text_naaclhlt}

\end{document}